\documentclass[letterpaper, 10 pt, conference]{ieeeconf}  

\IEEEoverridecommandlockouts               
\usepackage{amsmath,amssymb,subfig,hyperref,url,mathrsfs,array,graphicx,bm,balance}
\usepackage[T1]{fontenc}

\newcommand{\figref}[1]{Fig.~\ref{#1}}
\newcommand{\secref}[1]{Sec.~\ref{#1}}

\title{\LARGE \bf
Hierarchical Adaptive Control for Collaborative Manipulation of a Rigid Object by Quadrupedal Robots}

\author{Mohsen Sombolestan and Quan Nguyen
\thanks{M. Sombolestan and Q. Nguyen are with the Department of Aerospace and Mechanical Engineering, University of Southern California, Los Angeles, CA 90089, email: {\tt somboles@usc.edu, quann@usc.edu}.}}

\begin{document}

\maketitle
\begin{abstract}
Despite the potential benefits of collaborative robots, effective manipulation tasks with quadruped robots remain difficult to realize. In this paper, we propose a hierarchical control system that can handle real-world collaborative manipulation tasks, including uncertainties arising from object properties, shape, and terrain. Our approach consists of three levels of controllers. Firstly, an adaptive controller computes the required force and moment for object manipulation without prior knowledge of the object's properties and terrain. The computed force and moment are then optimally distributed between the team of quadruped robots using a Quadratic Programming (QP)-based controller. This QP-based controller optimizes each robot's contact point location with the object while satisfying constraints associated with robot-object contact. Finally, a decentralized loco-manipulation controller is designed for each robot to apply manipulation force while maintaining the robot's stability. We successfully validated our approach in a high-fidelity simulation environment where a team of quadruped robots manipulated an unknown object weighing up to 18 kg on different terrains while following the desired trajectory.
\end{abstract}
\section{Introduction} \label{sec: Introduction}
Legged robots are known for their ability to move quickly and maneuver easily due to their versatile locomotion skills. The advancement of model predictive control (MPC) for legged robots \cite{DiCarlo2018a, Li2021} has facilitated the creation of real-time control systems that can execute diverse walking gaits.
Most research on quadruped robots has concentrated on locomotion \cite{Focchi2017a, Bledt2018}, and loco-manipulation \cite{Chiu2022AManipulation, Sleiman2021AManipulation, Zimmermann2021GoArm, Rigo2022ContactControl, Wolfslag2020OptimisationRobots, Ferrolho2022RoLoMa:Arms} by a single robot. The approaches are even extended to the problem with significant uncertainty in robot model \cite{Sombolestan2023AdaptiveTerrain,Sombolestan2021} as well as manipulating an object with unknown property \cite{Sombolestan2023HierarchicalRobots}. However, a limited number of works explore collaboration among multiple quadruped robots. In scenarios with multiple general-purpose robots available rather than specialized, larger robots, collaboration among several quadruped robots can prove highly advantageous. 
The group of robots can work together to perform collaborative tasks beyond a single robot's capabilities, such as object manipulation in industrial factory locations and last-mile delivery operations.

The use of multiple quadruped robots for towing a load with cables towards a target while avoiding obstacles has been explored in \cite{Yang2022CollaborativeRobots}. However, in manipulation tasks, including the work mentioned above, the controller often necessitates prior knowledge of the manipulated object and terrain, such as the object's mass, geometry, and terrain friction coefficient.
Nonetheless, in many practical scenarios, the parameters of the manipulated object are generally unknown, especially if the object is non-geometric or asymmetric. Hence, the robot should be capable of adapting to a wide range of objects.

\begin{figure}[t!]
	\center
	\includegraphics[width=1\linewidth]{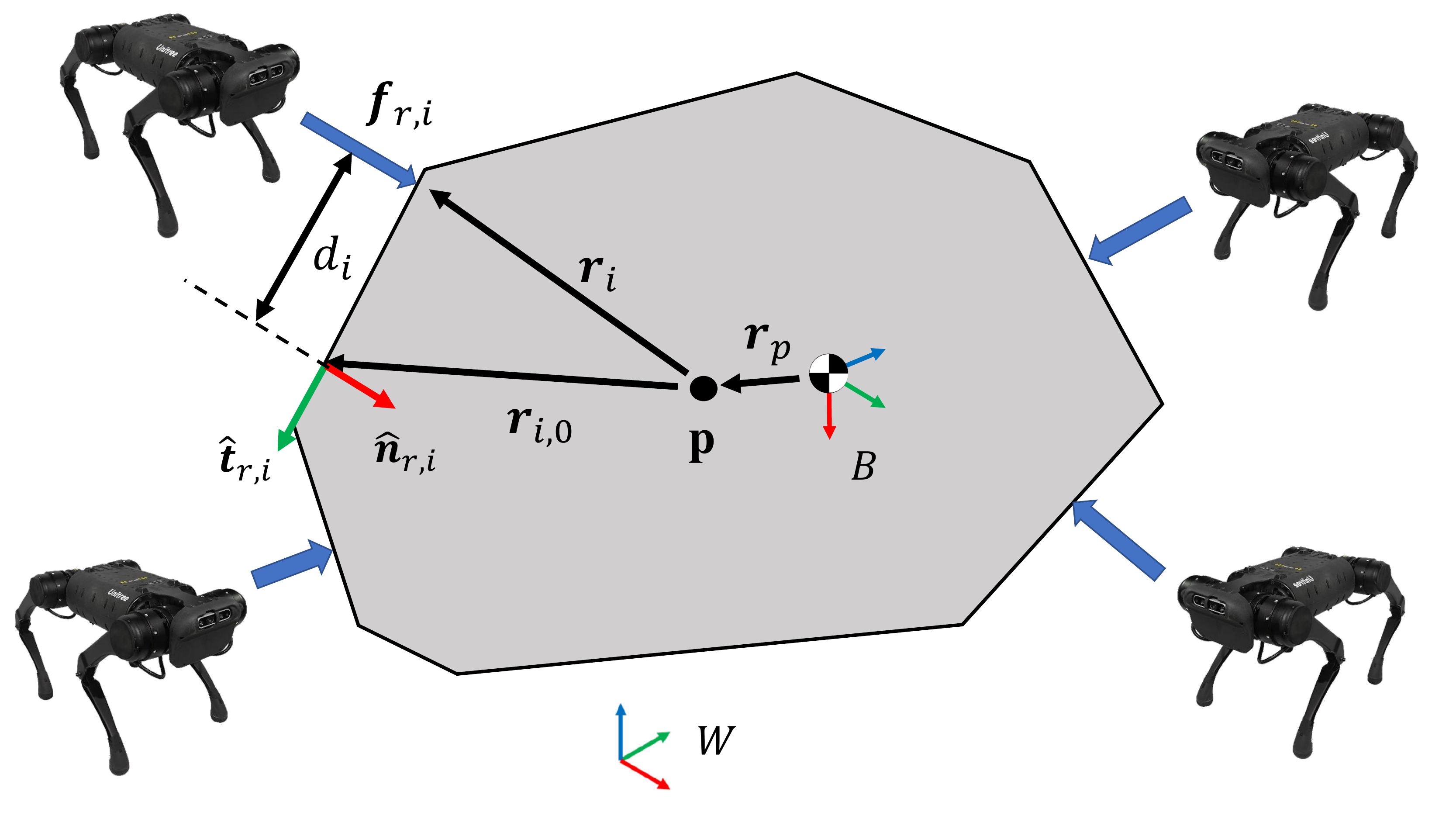}
	\caption{{\bfseries{Schematic of manipulation task.}} $\bm{p}$ is an arbitrary measurement reference point on the object. Object's properties such as mass, inertia, and COM location $\bm{r}_p$ are unknown. Simulation results: \protect\url{https://youtu.be/cHofdxolZk4}.}
	\label{fig: schematic}
	\vspace{-1em}
\end{figure}

Adaptive control has been employed in some prior research for collaborative manipulation in mobile robots without making assumptions about the object's mass. Both centralized controllers \cite{Hu1995MotionTasks, Li2008RobustManipulators} and decentralized controllers \cite{Liu1998DecentralizedCooperations, Verginis2017RobustInformation, Culbertson2021DecentralizedBodies, Fink2008Multi-robotObstacles} have been developed in this regard.
However, these approaches rely on the rigid connection between the object and robots during the manipulation task, which is problematic in quadrupedal robots application. The rigid connection can impact the robots' stability and limit the team of the robots' initial configuration before starting the manipulation task.
Additionally, in some instances, the measurement of the manipulators' relative positions from the center of mass (COM) is required \cite{Prattichizzo2008Grasping}. However, this assumption is impractical for non-geometric objects with an unknown COM location (see \figref{fig: schematic}). 
Our approach allows the robots to begin in a random location and then engage in object manipulation. In addition, there are no assumptions regarding object properties such as mass, inertia, and COM location, as well as terrain properties such as friction coefficient.

\begin{figure*}[t!]
	\center	\includegraphics[width=1\linewidth]{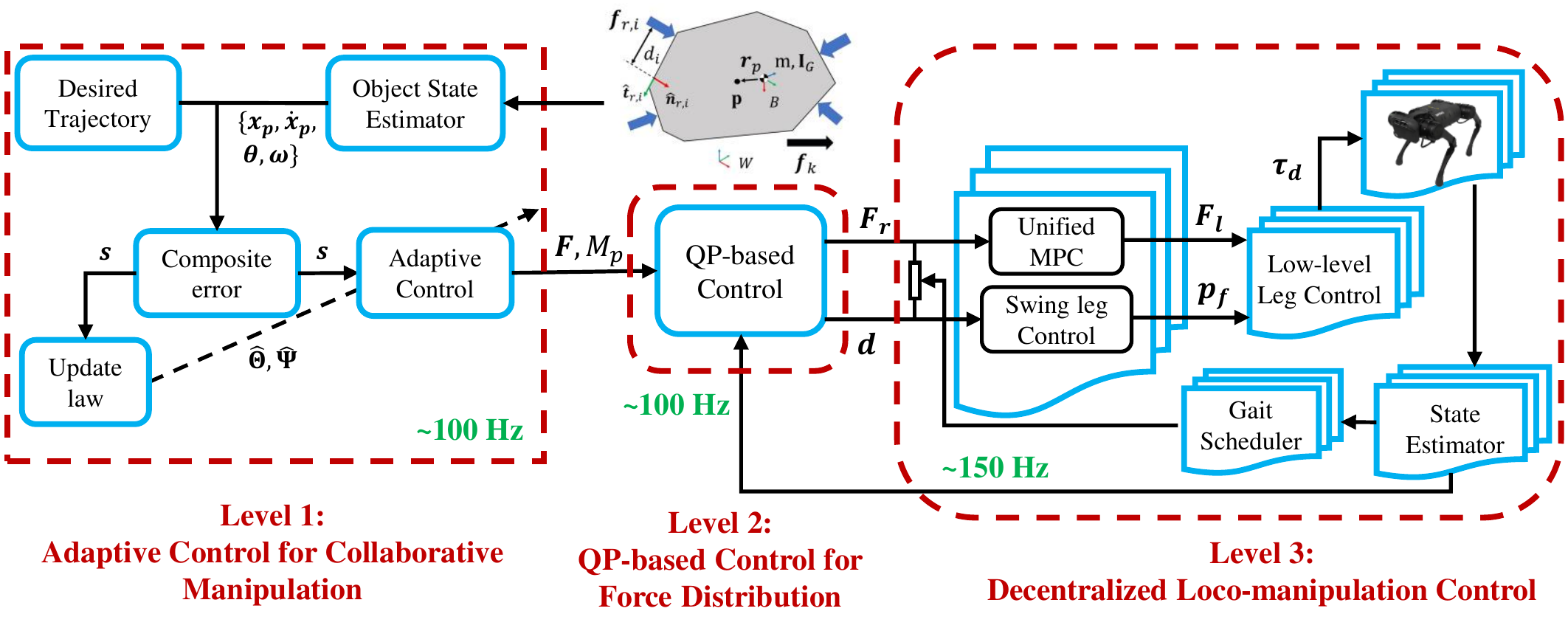}
	\caption{{\bfseries{Block diagram of our proposed approach}}. Our method contains three levels of controllers; each runs at a specific frequency.}
	\label{fig: block_diagram}
 \vspace{-1em}
\end{figure*}
This paper presents a hierarchical adaptive control for manipulating an unknown rigid object collaboratively using multiple quadruped robots.
In our previous work \cite{Sombolestan2023HierarchicalRobots}, we introduced a unified MPC framework that utilized robot locomotion to manipulate an object effectively without sacrificing robot balance. However, that work was limited to one-direction object manipulation with a single robot. 
In this paper, we aim to tackle the problem of planar collaborative manipulation of a heavy unknown object via multiple quadruped robots. 
Our proposed approach involves developing a controller that utilizes Quadratic Programming (QP) and is inspired by QP-based balancing control for quadruped robots \cite{Focchi2017a}. This controller is integrated with an adaptive controller to compensate for the object uncertainties and can optimally distribute manipulation force among multiple robots while adjusting each robot's contact point location.
The contribution of this paper is as follows:
\begin{itemize}
    \item We introduced a novel hierarchical control framework composed of three levels of controllers to facilitate the cooperative manipulation of an unknown asymmetric object using quadruped robots (see \figref{fig: block_diagram}).
    
    \item At the first level, we developed an adaptive controller that calculates the required force and moment for the object to follow the desired trajectory over time accurately. It should be noted that certain attributes of the object, such as mass, inertia, COM location, and frictional forces, are unknown.
    
    \item Next, we will propose a QP-based controller that will effectively allocate the manipulation force and moment among a group of quadruped robots while determining the optimal contact point for each robot. The QP controller is designed to meet the constraints related to the contact between the object and robots.
    
    \item In the last level, we utilize a decentralized loco-manipulation control for each robot that leverages the robot's locomotion to manipulate the object without losing the robot's stability.
    
    \item We validated the efficacy of our approach through high-fidelity simulations conducted on a team of Unitree A1 robots. We also compared our method and several baseline controllers to demonstrate the superiority of our framework. By employing our proposed approach, a team of robots can manipulate an unknown object weighing up to 18 kg across various terrains while accurately tracking the desired trajectory.
    
\end{itemize}

The rest of the paper is structured as follows: a brief overview of the control system is presented in \secref{sec: overview}. This is followed by a comprehensive explanation of our proposed method, which provides more details on the design of the three levels of our control system in \secref{sec: Method}. Furthermore, numerical validation is demonstrated in \secref{sec: simulation}. Finally, concluding remarks are provided in \secref{sec: conclusion}.

\section{Control System Overview} \label{sec: overview}

Our hierarchical proposed approach is illustrated in \figref{fig: block_diagram}. The control system has three levels according to \figref{fig: block_diagram}. We briefly introduce each level in this section, then in \secref{sec: Method}, we will discuss each part in more detail.

The first level of our approach involves designing an adaptive control system that enables the manipulated object to follow a desired trajectory. The object may have an asymmetric shape and unknown properties, and measurements of its state (such as position and velocity) will be taken relative to an arbitrary reference point, which need not be the COM. We assume that the object's mass, moment of inertia, and COM location are all unknown, as is the magnitude and direction of any external wrench (such as friction force) acting on the object. Using the adaptive controller, the control system can adapt to estimates of the object's properties.

Next, the force and moment values calculated by the adaptive control must be appropriately distributed among the robots. We developed a QP-based controller for this purpose, which enables each robot to apply an optimal force while adjusting its contact point location with respect to the object. \figref{fig: schematic} illustrates a schematic of multiple quadruped robots collaborating to manipulate an object. The robots are not rigidly attached to the object; they can only push the object in one direction while adjusting their position with respect to the object (as represented by $d_i$). The force direction of each robot ($\hat{\bm{n}}_{r,i}$) is treated as a constraint in the QP formulation. 

Lastly, a decentralized loco-manipulation control is created for each robot, considering the distributed force. The loco-manipulation controller consists of a unified MPC that incorporates both the locomotion controller for quadruped robots and the desired distributed manipulation force for each robot. This level serves as the critical control component for each robot and should run at a frequency of 150 Hz to ensure robust locomotion. In contrast, the higher-level parts of the control system need not be updated as frequently as the loco-manipulation controller. As such, the QP-based and adaptive controllers are run at a lower frequency of 100 Hz to free up sufficient processing units for the loco-manipulation control.

\section{Proposed Method} \label{sec: Method}
This section will elaborate on our hierarchical adaptive control for collaborative manipulation of an unknown asymmetric object using multiple quadruped robots. The object undergoes translational and rotational motion, led by a team of $n$ robots. A collaborative manipulation task is illustrated in \figref{fig: schematic}. First, we define our problem and the assumption that we will make to solve the problem, then we will introduce each level of our proposed approach illustrated in \figref{fig: block_diagram}. 

\subsection{Problem Definition}
There exists a world frame $W$ and a body frame $B$ attached to the object's center of mass (COM). There is a body-fixed point $\bm{p}$, which is the reference point for all the object's measurements and can be picked arbitrarily. As we indicated before, the properties of the object, such as mass ($m_b$), body-frame inertia about COM ($\bm{I}_G$), and COM position ($\bm{r}_p$) are unknown. Each robot starts at an initial point $\bm{r}_{i,0}$ from $\bm{p}$ on the object surface, and they have their own position estimation $\bm{r}_{i}$. 
A group comprising $n$ robots is expected to work together in manipulating the object. The number of robots involved in the task may vary as our approach enables the distribution of force among the active robots.
The robots can freely move along the object's surface, which means they are not rigidly connected to the object. Additionally, we assume that the friction between the robot and the object at the contact point is negligible. Therefore, each robot can only apply perpendicular force $\bm{f}_{r,i}$ to the object's surface (along $\hat{\bm{n}}_{r,i}$) while moving $d_i$ tangential to the object surface (along $\hat{\bm{t}}_{r,i}$). Note that all vectors defined in \figref{fig: schematic} such as $\bm{f}_{r,i}$, $\bm{r}_{i,0}$, $\bm{r}_p$, $\hat{\bm{n}}_{r,i}$, and $\hat{\bm{t}}_{r,i}$ are represented in the body-frame $B$.

Based on the above problem definition and assumptions made, we will describe the three levels of our control system shown in \figref{fig: block_diagram} in the following subsections.

\subsection{Adaptive Control for Object Manipulation}
\subsubsection{Equation of Motion for a Rigid Object}
The equation motion of a rigid body object can be written as follow:
\begin{align}
&\bm{F} = m_b {\bm{\ddot{x}}}_G
\\
&\bm{M}_G = \bm{R} \bm{I}_G \bm{R}^T \dot{\bm{\omega}} + \bm{\omega} \times (\bm{R} \bm{I}_G \bm{R}^T \bm{\omega})
\end{align}
where $\bm{R}$ is the rotation matrix from body frame $B$ to the world frame $W$, $\bm{\omega}$ is the angular velocity of the object, and $\dot{\bm{\omega}}$ is the angular acceleration. Since, in our problem, the COM position is unknown, we should derive the equation of motion of the rigid object with respect to the reference point $\bm{p}$:
\begin{align}
\label{eq: translation}
&\bm{F} = m_b \bm{\ddot{x}}_p - m_b(\dot{\bm{\omega}} \times \bm{R}\bm{r}_p) - m_b \bm{\omega} \times (\bm{\omega} \times \bm{R} \bm{r}_p) 
\\
 \label{eq: rotational}
&\bm{M}_p = \bm{R} \bm{I}_p \bm{R}^T \dot{\bm{\omega}} + \bm{\omega} \times (\bm{R} \bm{I}_p \bm{R}^T \bm{\omega}) - m_b \bm{R} \bm{r}_p \times \bm{\ddot{x}}_p
\end{align}
 where $\bm{F}$ and $\bm{M}_p$ are the force and moment required for object manipulation, respectively, $\ddot{\bm{x}}_p$ is the object's linear acceleration at point $\bm{p}$, and $\bm{I}_p$ is the object's moment of inertia with respect to $\bm{p}$.

Considering that the object is on the ground and will be manipulated within planar coordinates, we can limit our focus to the planar aspect of the equation of motion. To achieve this, we define the configuration variable $\bm{q} = [\bm{x}_p, \theta]$, where $\bm{x}_p$ is the position of reference point $\bm{p}$ in the world frame and $\theta$ represents the object's yaw angle. We also take into account an external wrench $\bm{f}_k$ and express the equation of motion in a compact form as follows:
 \begin{align}
 \label{eq: equation of motion}
\bm{\tau} = \bm{H}(\bm{q})\ddot{\bm{q}} + \bm{C}(\bm{q}, \bm{\dot{q}})\bm{\dot{q}}  + \bm{f}_k
\end{align}
where $\bm{\tau}$ is the wrench applied to the rigid object from a team of robots. By defining $\bm{r}_p = [r_x;r_y]$ and $I_{p,zz}$ as the moment of inertia about the normal direction to the ground, $\bm{H}(\bm{q})$ and $\bm{C}(\bm{q}, \bm{\dot{q}})$ can be represented as follow:
\begin{align} \label{eq: 2D motion}
    &\bm{H}(\bm{q}) = m_b \bm{R} \left[\begin{array}{ccc}
    1 & 0 & r_y \\
    0 & 1 & -r_x\\
    r_y & -r_x & \frac{I_{p,zz}}{m_b} \end{array} \right] \bm{R}^T
\\ \nonumber
    &\bm{C}(\bm{q}, \bm{\dot{q}})\bm{\dot{q}} = m_b \omega^2 \bm{R} \left[\begin{array}{c}
    r_x \\
    r_y\\
    0 \end{array} \right] 
\end{align}

\subsubsection{Adaptive Control} \label{sec: adaptive control}
In the adaptive control for manipulators \cite{Slotine1991AppliedControl}, it is common to use a linear combination of position and velocity error, denoted as $\bm{s}$. This approach results in exponentially stable dynamics when the surface $\bm{s}=0$ is reached. Hence, we define the composite error as follows:
\begin{align}
\bm{s} = \left[\begin{array}{c}
\bm{\dot{x}}_e + \lambda \bm{x}_e \\
\omega_e + \lambda \theta_e
\end{array} \right] 
\end{align}
where $\bm{x}_e$, $\bm{\dot{x}}_e$, $\theta_e$, and $\omega_e$ represented the tracking error for $\bm{x}_p$, $\bm{\dot{x}}_p$, $\theta$, and $\omega$, respectively. Then we define the reference velocity as follows:
\begin{align} \label{eq: reference velocity}
\bm{\dot{q}}_r = \bm{\dot{q}} - \bm{s}
\end{align}

The dynamic equation \eqref{eq: equation of motion} depends linearly on an unknown parameter vector $\bm{\Theta}$ \cite{Slotine1991AppliedControl}.
Thus, we can decompose the equation of motion into a known regressor matrix $\bm{Y}_{\Theta}$ and vector of unknown parameter $\bm{\Theta}$.
\begin{align} \label{eq: regressor1}
\bm{H}\ddot{\bm{q}}_r + \bm{C}\bm{\dot{q}}_r = \bm{Y}_{\Theta} \bm{\Theta}
\end{align}
We can exploit the same property for the unknown external wrench as well and define that in terms of known regressor matrix $\bm{Y}_{\Psi}$ and vector of an unknown parameter $\bm{\Psi}$.
\begin{align} \label{eq: regressor2}
\bm{f}_k = \bm{Y}_{\Psi} \bm{\Psi}
\end{align}

Now, we propose the control and adaption laws required to be applied to the object to track the desired trajectory asymptotically. The control law would be:
\begin{align} \label{eq: control law}
\bm{\tau} = \left[\begin{array}{c} \bm{F} \\ M_p \end{array} \right] = \bm{Y}_{\Theta} \hat{\bm{\Theta}} + \bm{Y}_{\Psi} \hat{\bm{\Psi}} - \bm{K}_D \bm{s} 
\end{align}
where $\hat{\bm{\Theta}}$ and $\hat{\bm{\Psi}}$ are estimated vector of unknown parameters and $\bm{K}_D$ is a positive definite matrix. The first two terms in control law \eqref{eq: control law} are related to the dynamic estimation, and the last term is a PD term which leads the object to follow the desired trajectory. Moreover, adaptation laws are proposed as follows:
\begin{align}
\label{eq: adaptation law 1}
\dot{\hat{\bm{\Theta}}} = -\bm{\Gamma}_\Theta  {\bm{Y}_\Theta}^T \bm{s} \\ \label{eq: adaptation law 2}
\dot{\hat{\bm{\Psi}}} = -\bm{\Gamma}_\Psi  {\bm{Y}_\Psi}^T \bm{s}
\end{align}
which $\bm{\Gamma}_\Theta$ and $\bm{\Gamma}_\Psi$ are positive definite matrices. In \secref{sec: stability}, we will provide a detailed explanation of the design of the control law \eqref{eq: control law} and adaptation laws \eqref{eq: adaptation law 1}, \eqref{eq: adaptation law 2}, as well as the stability proof. 
Note that we employ direct adaptive control, and it is not our expectation for the estimated vector of unknown parameters to converge to the actual value.

\vspace{7pt}
\subsubsection{Stability Proof} \label{sec: stability}
Let us consider the following Lyapunov function:
\begin{align}
\label{eq: lyapunov function}
V(t) = \frac{1}{2}(\bm{s}^T \bm{H} \bm{s} + \tilde{\bm{\Theta}}^{T} {\bm{\Gamma}_\Theta}^{-1} \tilde{\bm{\Theta}} + \tilde{\bm{\Psi}}^{T} {\bm{\Gamma}_\Psi}^{-1} \tilde{\bm{\Psi}} )
\end{align}
where $\tilde{\bm{\Theta}} = \hat{\bm{\Theta}} - \bm{\Theta}$ and $\tilde{\bm{\Psi}} = \hat{\bm{\Psi}} - \bm{\Psi}$ are vectors of estimation error. Note that according to the definition of $\bm{H}$ in \eqref{eq: 2D motion}, $\bm{H}$ is a positive definite matrix.
Since $\bm{\Theta}$ and $\bm{\Psi}$ are constant vectors, the estimation error derivative $\dot{\tilde{\bm{\Theta}}}$, $\dot{\tilde{\bm{\Psi}}}$ are the same as the estimation derivative $\dot{\hat{\bm{\Theta}}}$, $\dot{\hat{\bm{\Psi}}}$. By considering this property, we will take the derivative of $V(t)$: 
\begin{align} \label{eq: lyapunov function differential}
\dot{V}(t) = \bm{s}^T \bm{H} \dot{\bm{s}} + \frac{1}{2}\bm{s}^T \dot{\bm{H}} \bm{s} + {\tilde{\bm{\Theta}}}^{T} {\bm{\Gamma}_\Theta}^{-1} \dot{\hat{\bm{\Theta}}} + {\tilde{\bm{\Psi}}}^{T} {\bm{\Gamma}_\Psi}^{-1} \dot{\hat{\bm{\Psi}}}.
\end{align}
According to the definition of reference velocity in \eqref{eq: reference velocity}, we know $\dot{\bm{q}} = \bm{s} + \dot{\bm{q}}_r$ and $\dot{\bm{s}}= \ddot{\bm{q}} - \ddot{\bm{q}}_r$. Therefore, by considering the equation of motion \eqref{eq: equation of motion}, the first term in equation \eqref{eq: lyapunov function differential} can be expanded as follows:
\begin{align}
\label{eq: expand}
\bm{s}^T \bm{H} \dot{\bm{s}} &= \bm{s}^T \bm{H} (\ddot{\bm{q}} - \ddot{\bm{q}}_r) \\ \nonumber
    &= -\bm{s}^T \bm{C}\bm{s} + \bm{s}^T[\bm{\tau} - (\bm{H}\ddot{\bm{q}}_r + \bm{C}\dot{\bm{q}}_r) - \bm{f}_k ]
\end{align}
By using the property described in \eqref{eq: regressor1} and \eqref{eq: regressor2}, we substitute the \eqref{eq: expand} into \eqref{eq: lyapunov function differential}, then we have:
\begin{align} \label{eq: lyapunov function differential 2}
\dot{V}(t) =  \bm{s}^T[\bm{\tau} - \bm{Y}_{\Theta} \bm{\Theta} - \bm{Y}_{\Psi} \bm{\Psi}] + \frac{1}{2}\bm{s}^T (\dot{\bm{H}}-2\bm{C}) \bm{s} + \\ \nonumber
{\tilde{\bm{\Theta}}}^{T} {\bm{\Gamma}_\Theta}^{-1} \dot{\hat{\bm{\Theta}}} + {\tilde{\bm{\Psi}}}^{T} {\bm{\Gamma}_\Psi}^{-1} \dot{\hat{\bm{\Psi}}}.
\end{align}
The $\dot{\bm{H}}-2\bm{C}$ is a skew-symmetric matrix \cite{Culbertson2021DecentralizedBodies}, so the second term in \eqref{eq: lyapunov function differential 2} is zero. Finally, substituting the control law \eqref{eq: control law} and adaptation laws \eqref{eq: adaptation law 1}, \eqref{eq: adaptation law 2} into equation \eqref{eq: lyapunov function differential 2} yields:
\begin{align} \label{eq: lyapunov function differential 3}
\dot{V}(t) & =  - \bm{s}^T \bm{K}_D \bm{s} \leq 0.
\end{align}
According to the Lyapunov theorem \cite{Slotine1991AppliedControl}, the system is uniformly stable because $V(t)$ is positive definite and decrescent, and $\dot{V}(t)$ is negative semi-definite. As a result, the variables $\bm{s}$, $\tilde{\bm{\Theta}}$, and $\tilde{\bm{\Psi}}$ will remain bounded.

The expression in \eqref{eq: lyapunov function differential 3} indicates that $V(t)$ has a finite limit, and it can be easily demonstrated that $\dot{\bm{s}}$ is bounded \cite{Culbertson2021DecentralizedBodies}. As a result, $\ddot{V}(t)$ is bounded, as can be observed from the expression $\ddot{V}(t) = -2 \bm{s}^T \bm{K}_D \dot{\bm{s}}$. Since $\ddot{V}(t)$ is bounded, and $\dot{V}(t)$ is uniformly continuous in time, and $V(t)$ is lower bounded, the second version of Barbalat's Lemma \cite{Slotine1991AppliedControl} implies that $\dot{V}(t) \rightarrow 0$ as $t \rightarrow \infty$. Therefore, $\bm{s}$ also approaches zero as $t \rightarrow \infty$. When $\bm{s} = 0$, it can be shown that $\dot{\bm{x}}_e = - \lambda \bm{x}_e$ and $\omega_e = - \lambda \theta_e$, which corresponds to an asymptotically stable system.

\subsection{QP-based Control for Force Distribution} \label{sec: QP}
The adaptive control presented in \secref{sec: adaptive control} calculates the force $\bm{F}$ and moment $M_p$ required for object manipulation that the object's pose $\bm{q}$ track the desired pose $\bm{q}_d$. Since our approach, in general, is not limited to a specific number of robots, we need an optimal framework to distribute the manipulation force into each robot. Importantly, each agent has a constraint on the direction of the force that it can apply. Each robot starts from a random initial position $\bm{r}_{i,0}$ with respect to point $\bm{p}$ (see \figref{fig: schematic}). 
Then, the robot can only apply force perpendicular to the object's surface (along $\hat{\bm{n}}_{r,i}$) while moving $d_i$ tangential to the object's surface (along $\hat{\bm{t}}_{r,i}$) within a specific range on the object that allows the robot to navigate. To this end, we developed the following QP formulation to compute the optimal control input for each robot while satisfying constraints:
\begin{align} \label{eq: distributed QP}
\nonumber
\left[\begin{array}{c} {\bm{F}_r}^* \\ \bm{d}^*
\end{array} \right]=  \underset{\bm{F}_r ,\bm{d} \in \mathbb{R}^{n}}{\operatorname{argmin}}   \:\:   & \gamma_1 \| \bm{F}_r \|^2 + \gamma_2 \| \bm{F}_r - \bm{F}_{r,\textrm{prev}}^* \|^2 \\  & +\gamma_3 \| \bm{d} - \bm{d_{\textrm{prev}}}^* \|^2 \\
\mbox{s.t. } \quad \:\: & \textrm{(1):}~ \sum_{i = 1 }^{n} \bm{f}_{r,i} = \bm{R} \bm{F} \nonumber\\
\quad \:\: & \textrm{(2):}~ \sum_{i = 1 }^{n} \bm{r}_i \times \bm{f}_{r,i} = M_p   \nonumber\\
\quad \:\: & \textrm{(3):}~ F_{r,i} = 
\begin{cases}
    \geq 0 & \text{active}\\
    0      & \text{otherwise}
\end{cases}  \nonumber \\ 
\quad \:\: & \textrm{(4):}~ \underline{\bm{d}} \leq \bm{d} \leq \bar{\bm{d}} \nonumber \\
\mbox{\text{with:}} \quad \:\: & \nonumber \\ 
&\bm{f}_{r,i} = F_{r,i} \hat{\bm{n}}_{r,i} \nonumber \\
&\bm{r}_i = \bm{r}_{i,0} + d_i {\hat{\bm{t}}_{r,i}} \nonumber
\end{align}
where $\bm{F}_r$ is the vector of agents force magnitude ($\bm{F}_r = [F_{r,1}, F_{r,2}, \dots , F_{r,n}]^T \in \mathbb{R}^{n}$) and $\bm{d}$ is the vector of agents position on the object's surface ($\bm{d} = [d_1, d_2, \dots , d_n]^T \in \mathbb{R}^{n}$). 
The cost function contains three terms to minimize the force magnitude $\bm{F}_r$ as well as the change of the current solution with respect to the solution from the previous time-step for both force magnitude $\bm{F}_r$ and distance $\bm{d}$. 

The first two constraints are regarding achieving the desired manipulation force computed with adaptive control using our team of robots. Note that the calculated force $\bm{F}$ is represented in the fixed world frame $W$. Since all the agents' forces $\bm{f}_{r,i}$ are described in the object body frame $B$, we transform the manipulation force vector into the object body frame using rotation matrix $\bm{R}$. This fact does not affect the $M_p$ since, for a planar problem, we only have a moment about the direction normal to the ground, and $M_p$ is scalar. The third constraint is associated with agents' force. Since each active agent can only push the object forward, the force magnitude is always positive. If the robot is not in contact with the object, no force will be distributed to that robot. Finally, the last constraint ensures the robot will not exceed the surface limitation.

\subsection{Decentralized Loco-manipulation Control via Unified MPC}
This subsection will introduce a decentralized loco-manipulation control for each agent. We previously developed a unified MPC that considers both locomotion and manipulation for robots \cite{Sombolestan2023HierarchicalRobots}. The unified MPC regulates the manipulation force achieved in \secref{sec: QP} while maintaining the robot balance. First, we write the robot's equation of motion with the manipulation force based on the state representation presented in \cite{DiCarlo2018a}:
\begin{align}\label{eq: loco-manipulation dynamic}
 \bm{\dot{X}}_i = \bm{D}_i \bm{X}_i + \bm{G}_i \bm{F}_{l,i} + \bm{f}_{r,i}^{w} / m_i
\end{align}
where $\bm{f}_{r,i}^{w}$ is the force vector $\bm{f}_{r,i}$ represented in the world frame $W$, $m_i$ is the robot mass, $\bm{F}_{l,i}$ is the vector of ground reaction forces for all the legs, and $\bm{X}_i$ contains the robot's body's COM location, Euler angle, and velocities. More details on the equation as well as the definition of $\bm{D}$ and $\bm{G}$ can be found in \cite{Sombolestan2023HierarchicalRobots}. Note that the $i \in \{1, \dots, n\}$ represents the agent index number.

MPC employs linear discrete-time dynamics to predict the system's behavior over a finite time horizon. However, using a traditional discretization technique like zero-order hold requires incorporating the manipulation term $\bm{f}_{r,i}^{w}$ from equation \eqref{eq: loco-manipulation dynamic} into the state vector to create an extended vector for MPC formulation. As a result, equation \eqref{eq: loco-manipulation dynamic} can be rewritten as:
\begin{align}
\label{eq: combined SS}
\bm{\dot{\eta}}_i = \bar{\bm{D}}_i \bm{\eta}_i + \bar{\bm{G}}_i \bm{F}_{l,i} 
\end{align}
where 
\begin{align} \label{eq: extended SS components}
&\bm{\eta}_i = \left[\begin{array}{c} 
    \bm{X}_i \\ 
    \hline
    \bm{f}_{r,i}^{w} /m_i
    \end{array} \right] \in \mathbb{R}^{15}
\\ \nonumber
&\bar{\bm{D}}_i = \left[\begin{array}{@{}c|c@{}} 
        \begin{matrix}
           \bm{D}_{i} \in \mathbb{R}^{13 \times 13}
        \end{matrix}
        & \begin{matrix}
            \bm{0}_{6 \times 2} \\
            \bm{I}_{2 \times 2} \\
            \bm{0}_{5 \times 2} \\
        \end{matrix}
    \\
    \hline
    \bm{0}_{2 \times 13} & \bm{0}_{2 \times 2}
\end{array} \right] \in \mathbb{R}^{15 \times 15}
\\ \nonumber
&\bar{\bm{G}}_i = \left[\begin{array}{c}
    \bm{G}_i \\
    \hline
    \bm{0}_{2 \times 12}
\end{array} \right] \in \mathbb{R}^{15 \times 12}
\end{align}
where $\bm{\eta}_i$ is the augmented vector. Therefore, a linear MPC can be designed as follows:
\begin{align}
\label{eq: mpc}
\min_{\bm{F}_{l,i}} \quad & \sum_{j=0}^{k-1} {\tilde{\bm{X}}_{i,j+1}}^T \bm{Q} {\tilde{\bm{X}}_{i,j+1}} + {\bm{F}_{l,i,j}}^T \bm{P}_{i,j} \bm{F}_{l,i,j} \\ \nonumber
\textrm{s.t.} \quad & \tilde{\bm{X}}_{i,j+1} = \bm{X}_{i,j+1} - \bm{X}_{d,i,j+1} \\ \nonumber
& \bm{\eta}_{i,j+1} = \bar{\bm{D}}_{t,j} \bm{\eta}_{j} + \bar{\bm{G}}_{t,j} \bm{F}_{l,i,j} \\ \nonumber
& \underline{\bm{c}}_f \leq \bm{C}_f \bm{F}_{l,i,j} \leq \bar{\bm{c}}_f
\end{align}
where $k$ is the number of horizons, $\bm{X}_{d,i,j}$ is the robot desired state at time step $j$, $\bm{Q}$ and $\bm{P}$ are diagonal positive semi-definite matrices, $\bar{\bm{D}}_{t,j}$ and $\bar{\bm{G}}_{t,j}$ are discrete-time system dynamics matrices, and $\underline{\bm{c}}_f \leq \bm{C}_f \bm{F}_{l,i,j} \leq \bar{\bm{c}}_f$ represents friction cone constraints defined in \cite{Focchi2017a}. Note that the computed $d_i$ from the QP-based controller affects the robot's desired state $\bm{X}_{d,i,j}$.
\section{Numerical Simulation} \label{sec: simulation}
The purpose of this section is to demonstrate the effectiveness of our proposed approach through numerical simulations. Our simulations were conducted in a high-fidelity environment called Gazebo 11, with controllers implemented in ROS Noetic. The simulations involved a team of Unitree A1 robots attempting to manipulate an unknown asymmetric object to track its desired pose $\bm{q}_d$ despite uncertainty in both the object and terrain.
During the manipulation task, each robot stays in contact with one of the object's surfaces and adjusts its orientation accordingly, in alignment with the object's orientation.
We carried out multiple simulations with varying scenarios to demonstrate the team's adaptability. More details of our conducted simulations are shown in the supplemental video\footnote{\protect\url{https://youtu.be/cHofdxolZk4}}.

\subsection{Comparative Analysis}
We conducted comparative simulations to assess the efficacy of our proposed method. Specifically, we focused on evaluating the impact of the first two levels of controllers in our control system. During each phase of the evaluation, we analyzed the performance of the control system both with and without adaptive control and QP-based control.

\vspace{3pt}
\subsubsection{Effect of Adaptive Control}
\begin{figure}[t!]
	\centering
    \subfloat[snapshot of manipulating a total of 11 kg load]{\includegraphics[width=0.9\linewidth]{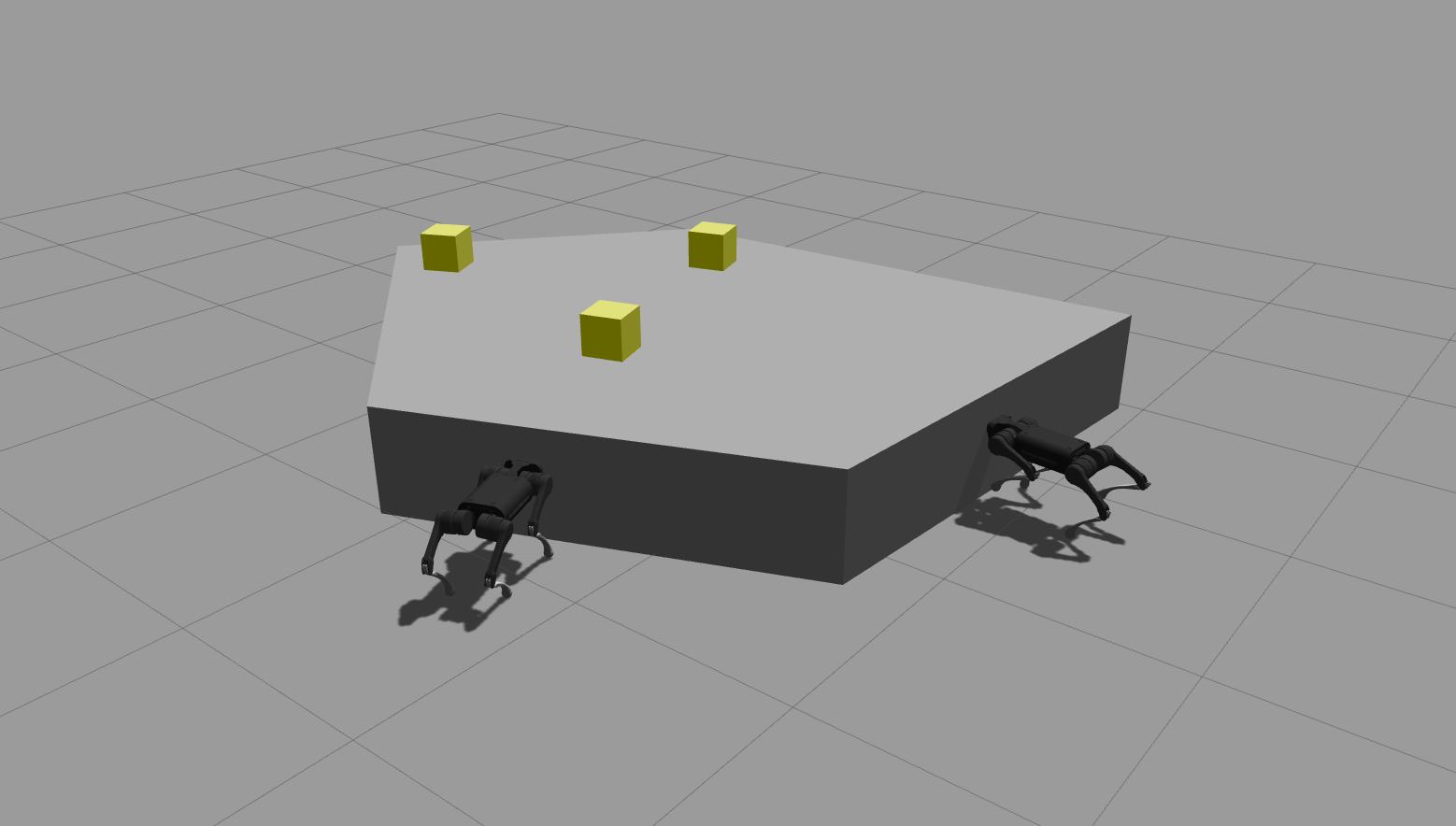}}
    \hfill
    \subfloat[Position error]{\includegraphics[width=1\linewidth]{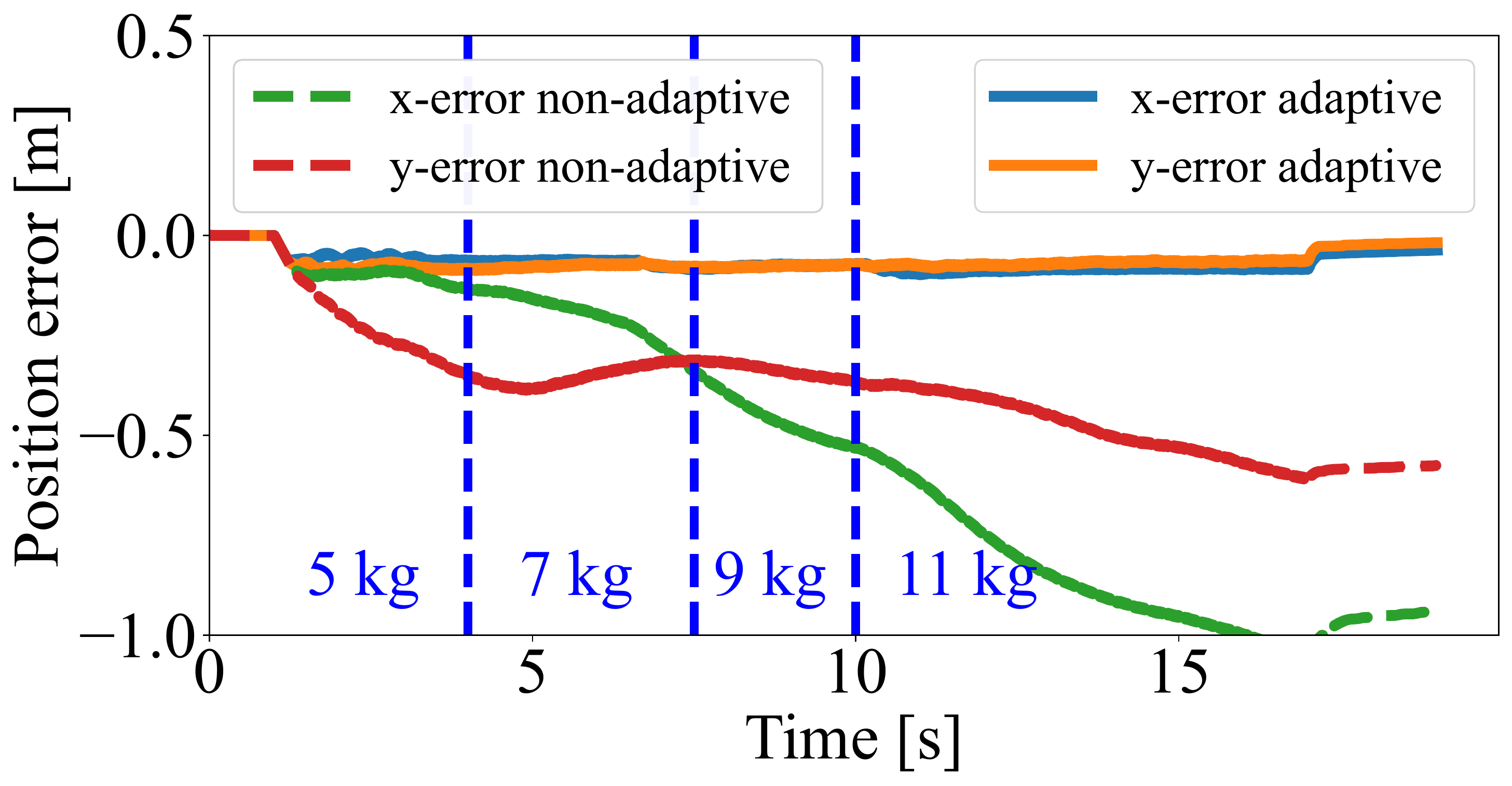}}
    \hfill
    \subfloat[Yaw error] {\includegraphics[width=1\linewidth]{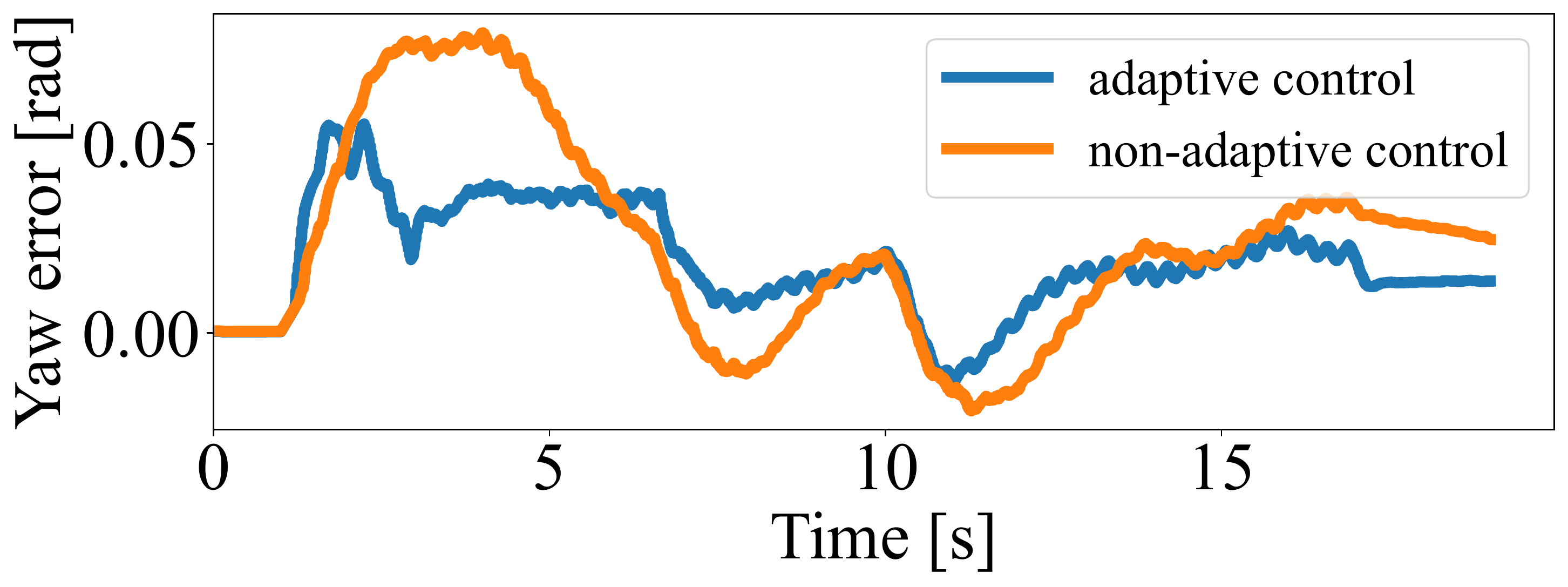}\label{subfig: adaptive_yaw}}

	\caption{{\bfseries{Effect of Adaptive Control}}. In the plots, we compare the results using adaptive and non-adaptive controllers. The team starts with an unknown 5 kg object, then three 2 kg loads will be dropped on top of the object.}
	\label{fig: compare_adaptive}
		\vspace{-1em}
\end{figure}

We compared our proposed approach and an alternative one that utilizes a PD controller at the first level instead of the adaptive controller. During the simulation, a team of robots manipulates an unknown asymmetric object weighing 5 kg, and three unknown objects weighing 2 kg each are randomly dropped onto the main object. The result for adaptive and non-adaptive methods are compared in \figref{fig: compare_adaptive}.

Using our proposed controller, the object successfully tracks the desired trajectory with minimal error and arrives at the target position with the intended orientation. However, the non-adaptive method fails to reach the target position within the specified time. It should be noted that the yaw tracking for both methods is almost the same (as seen in \figref{subfig: adaptive_yaw}), indicating that the team of robots can adjust the object's orientation using non-adaptive control, but they are unable to apply enough force for position tracking.

\vspace{3pt}
\subsubsection{Effect of QP-based Control}
\begin{figure}[t!]
	\centering
	\subfloat[Using QP-based control]{\includegraphics[width=0.48\linewidth]{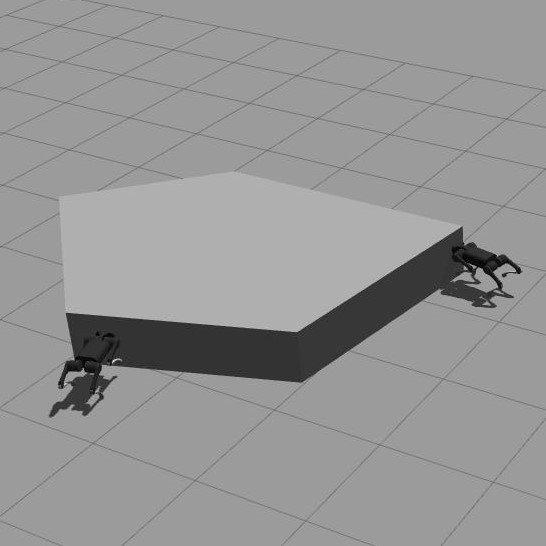}}
	\hfill
	\subfloat[Using heuristic method]{\includegraphics[width=0.48\linewidth]{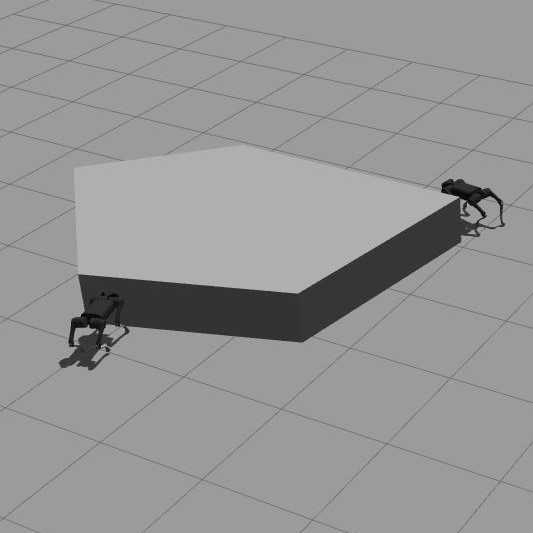}\label{subfig: lose contact}}
    \hfill
    \subfloat[Object trajectory]{\includegraphics[width=0.8\linewidth]{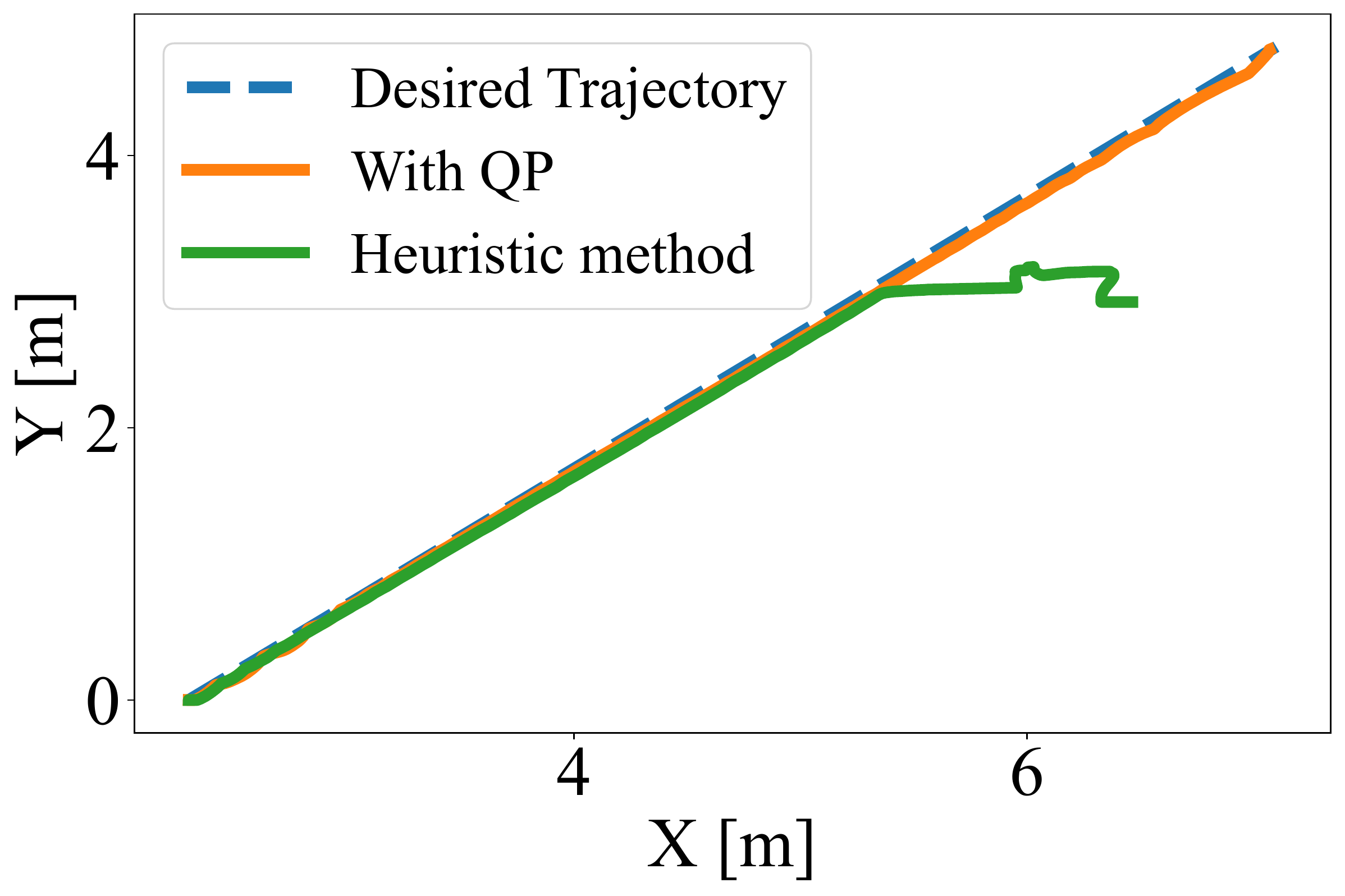}}
    \hfill
    \subfloat[Yaw tracking]{\includegraphics[width=1\linewidth]{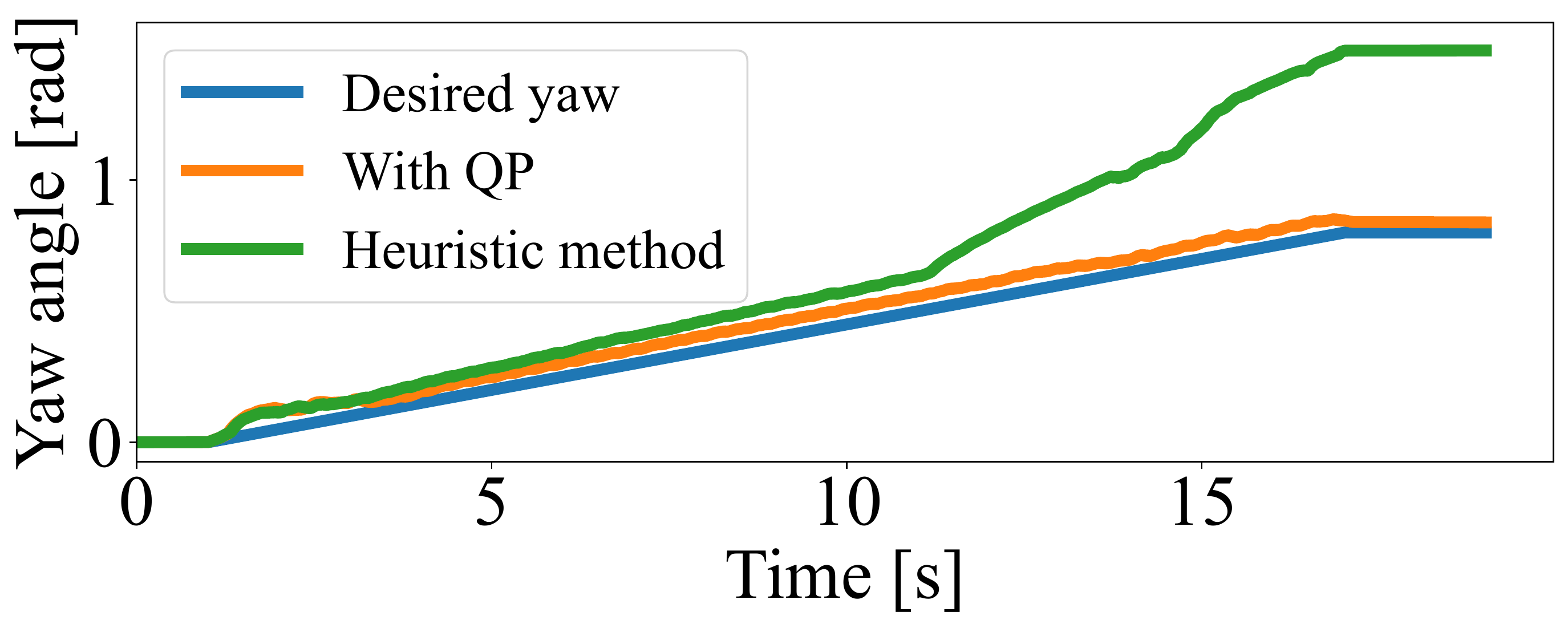}}

	\caption{{\bfseries{Effect of QP-based Control}}. A team of two robots tries to manipulate an unknown 5 kg object. The plots compare the result using a QP-based controller and a heuristic policy in the control system.}
	\label{fig: compare_QP}
		\vspace{-1em}
\end{figure}

We proceeded to investigate the effectiveness of the QP-based controller in our proposed method. To highlight the advantages of this controller, we utilized a heuristic approach to adjust the contact location $d_i$ based on the object's yaw angle error:
\begin{align}
\label{eq: heuristic}
d_i = k_p^d (\theta_d - \theta) 
\end{align}
We compared the performance of our method with the heuristic approach. The scenario involved a team of two robots attempting to manipulate an object in a straight line while adjusting its yaw angle. The comparison results are displayed in \figref{fig: compare_QP}.

As depicted in \figref{fig: compare_QP}, our proposed method, with the QP-based controller, can accurately track the desired trajectory. In contrast, when using the heuristic policy, one of the robots attempts to adjust its contact point to align with the desired yaw angle but exceeds the object surface limitation. During manipulation, the robot loses contact with the object (as shown in \figref{subfig: lose contact}), leading to a significant deviation from the intended trajectory. 

\subsection{Terrain Uncertainty}
\begin{figure}[t!]
	\centering
	\subfloat[Hardwood ground]{\includegraphics[width=0.48\linewidth]{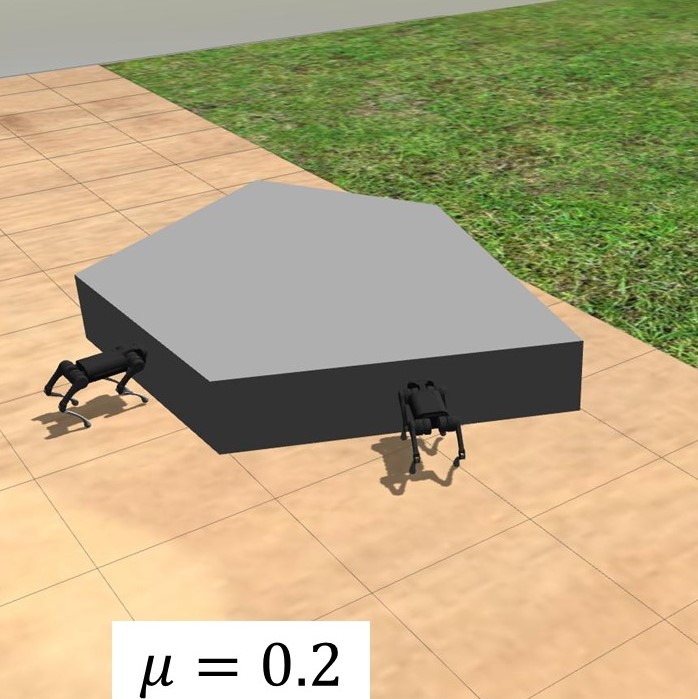}}
	\hfill
	\subfloat[Grass field]{\includegraphics[width=0.48\linewidth]{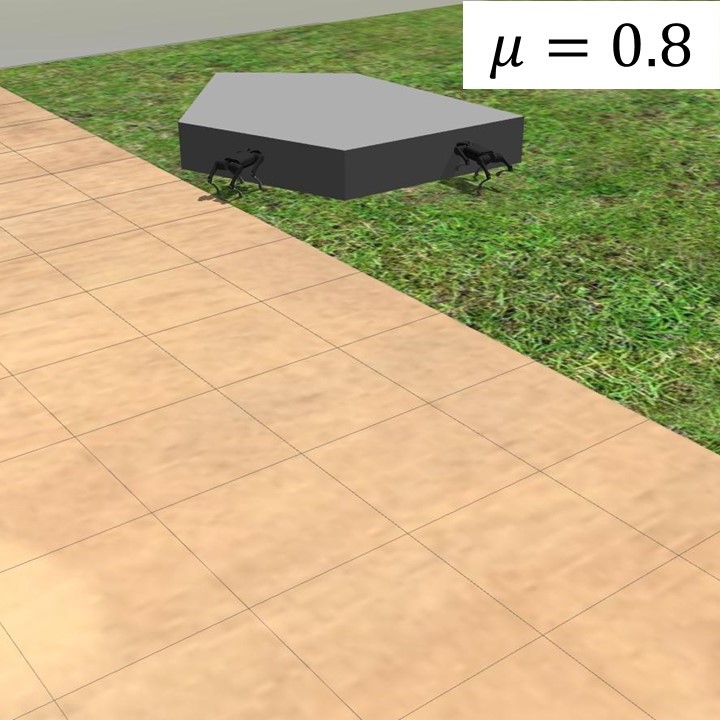}}
    \hfill
    \subfloat[Object trajectory]{\includegraphics[width=0.8\linewidth]{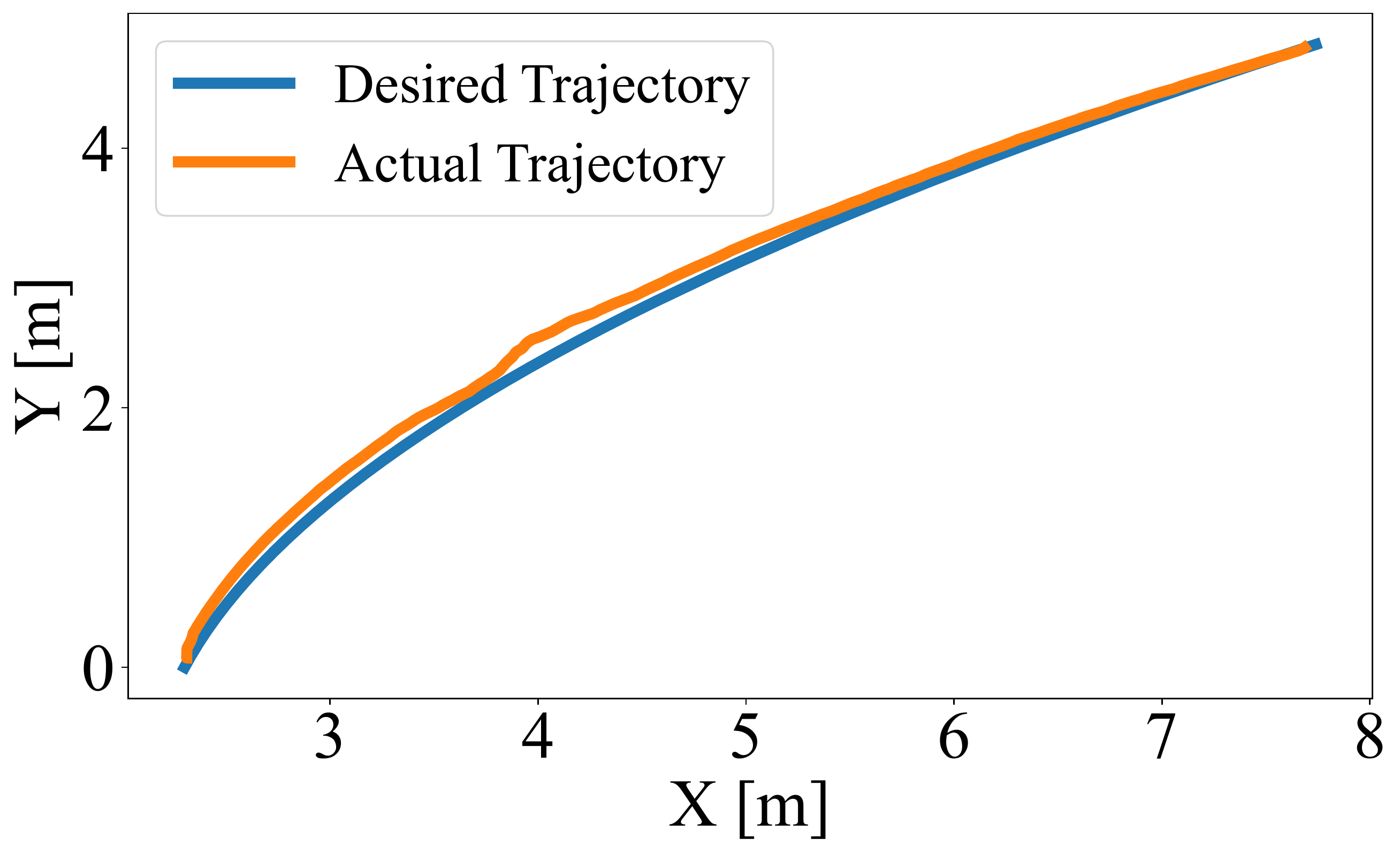}}
    \hfill
    \subfloat[Position error]{\includegraphics[width=1\linewidth]{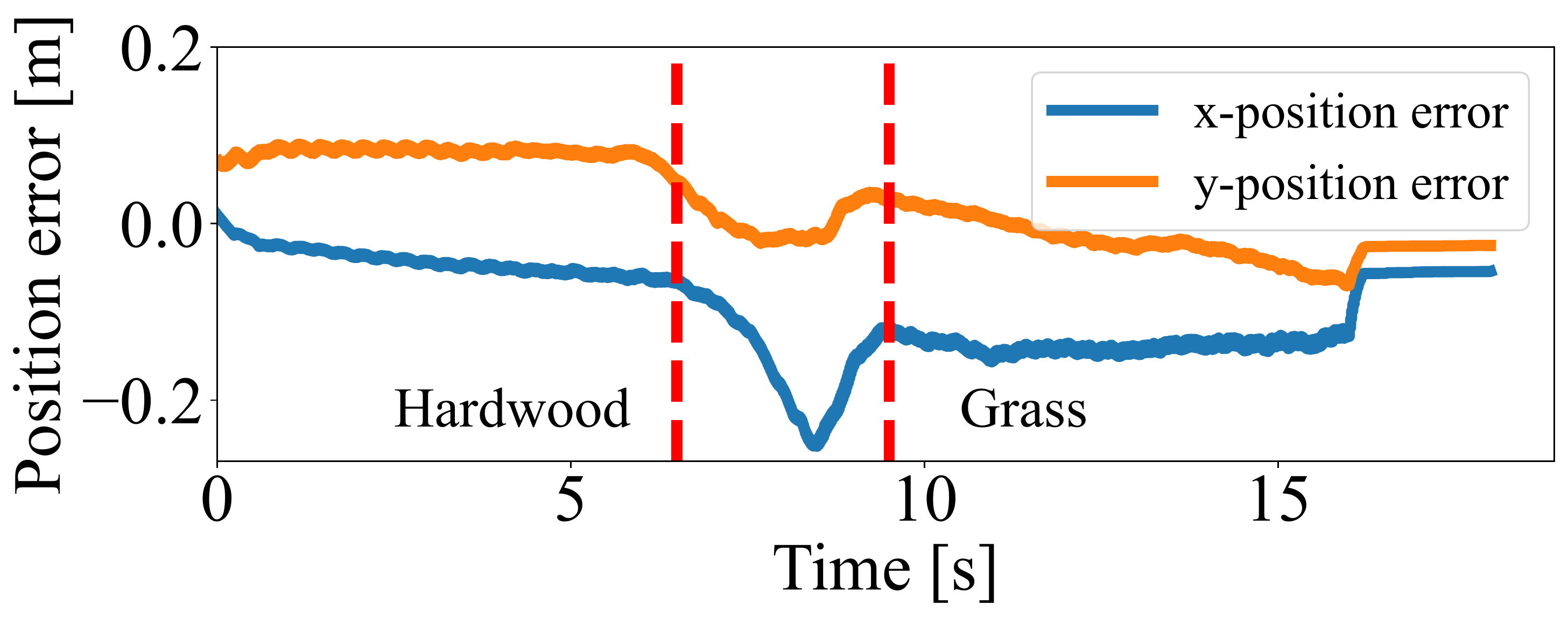}}
    \hfill
    \subfloat[Yaw tracking]{\includegraphics[width=1\linewidth]{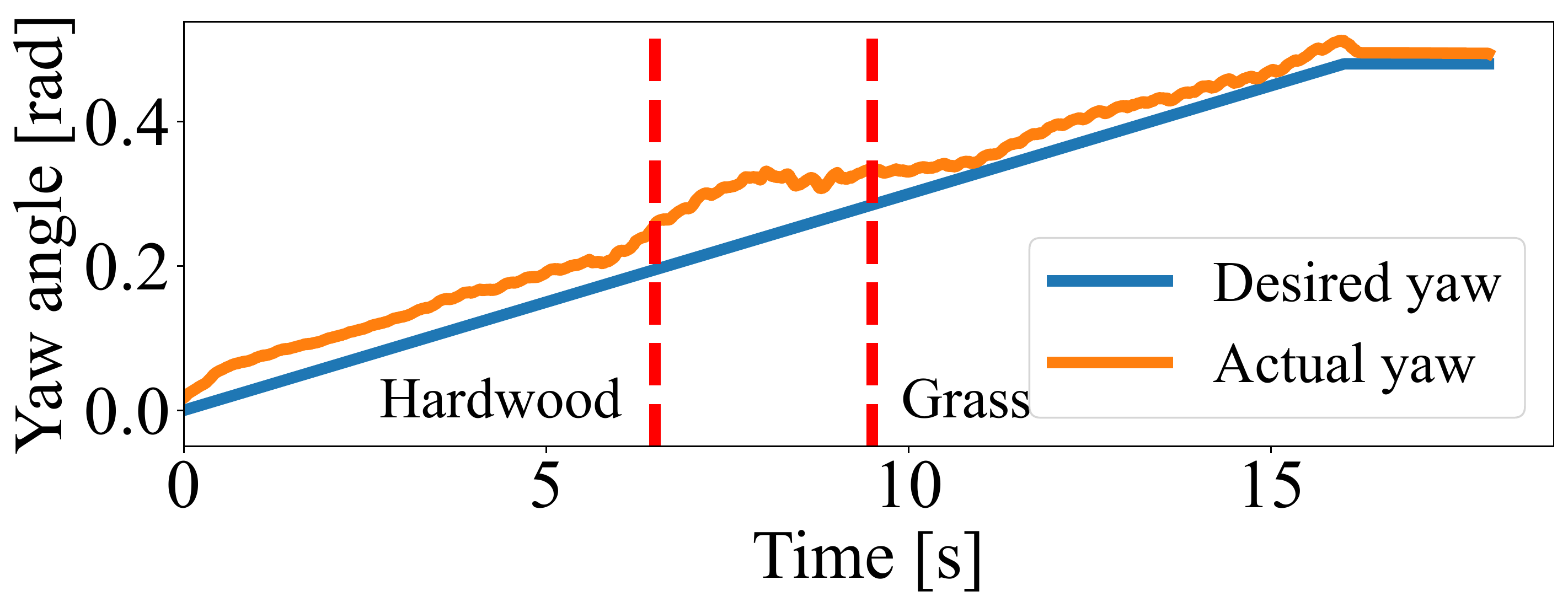}}	\caption{{\bfseries{Navigating surfaces with different friction properties}}. The transition part between two red dotted lines indicates when the object is on the grass, but the robot's feet are still on the hardwood ground.}
	\label{fig: various_friction}
		\vspace{-1em}
\end{figure}

Our next objective is to evaluate the robot's ability to adjust to terrain uncertainties. To achieve this, we will create a simulation where the team of robots moves through diverse terrains with varying friction properties on a desired curve trajectory while manipulating an unknown object weighing 5 kg. The robot will begin by navigating on a hardwood surface that has a friction coefficient of $\mu = 0.3$. Subsequently, it will traverse a grassy field that has a friction coefficient of $\mu = 0.8$. The results are presented in \figref{fig: various_friction}. As depicted in \figref{fig: various_friction}, the tracking error of the robot increases as it moves from hardwood ground to grass. During this transition, the object is partially on the grass, requiring more significant force for manipulation. However, the robot's feet are still on the hardwood ground, which has low friction, preventing the robot from exerting sufficient force for object manipulation.

\subsection{Collaborative Manipulation of a Heavy Load}
\begin{figure}[t!]
	\centering
	\subfloat[Before joining the third robot]{\includegraphics[width=0.48\linewidth]{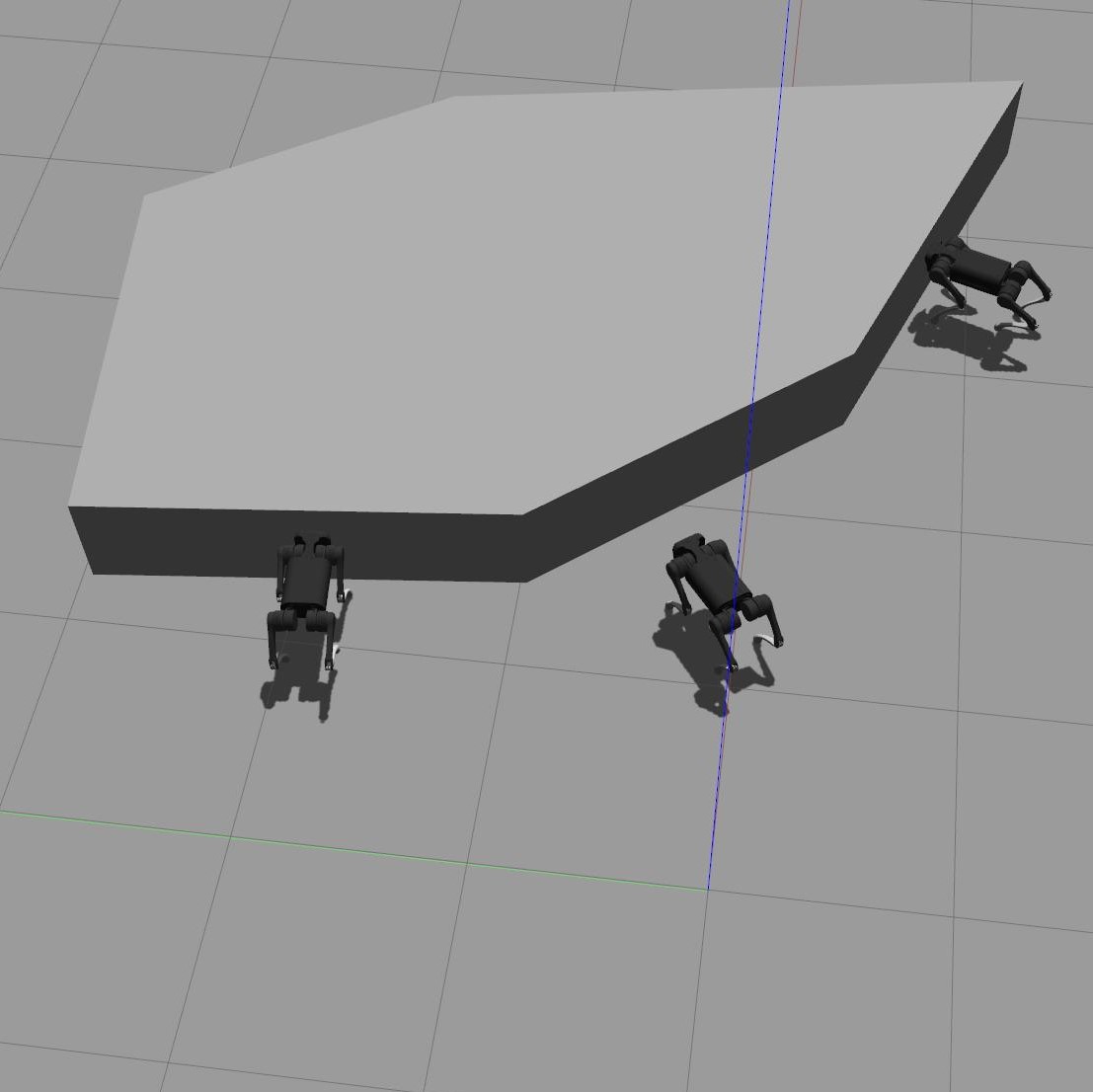}}
	\hfill
	\subfloat[After joining the third robot]{\includegraphics[width=0.48\linewidth]{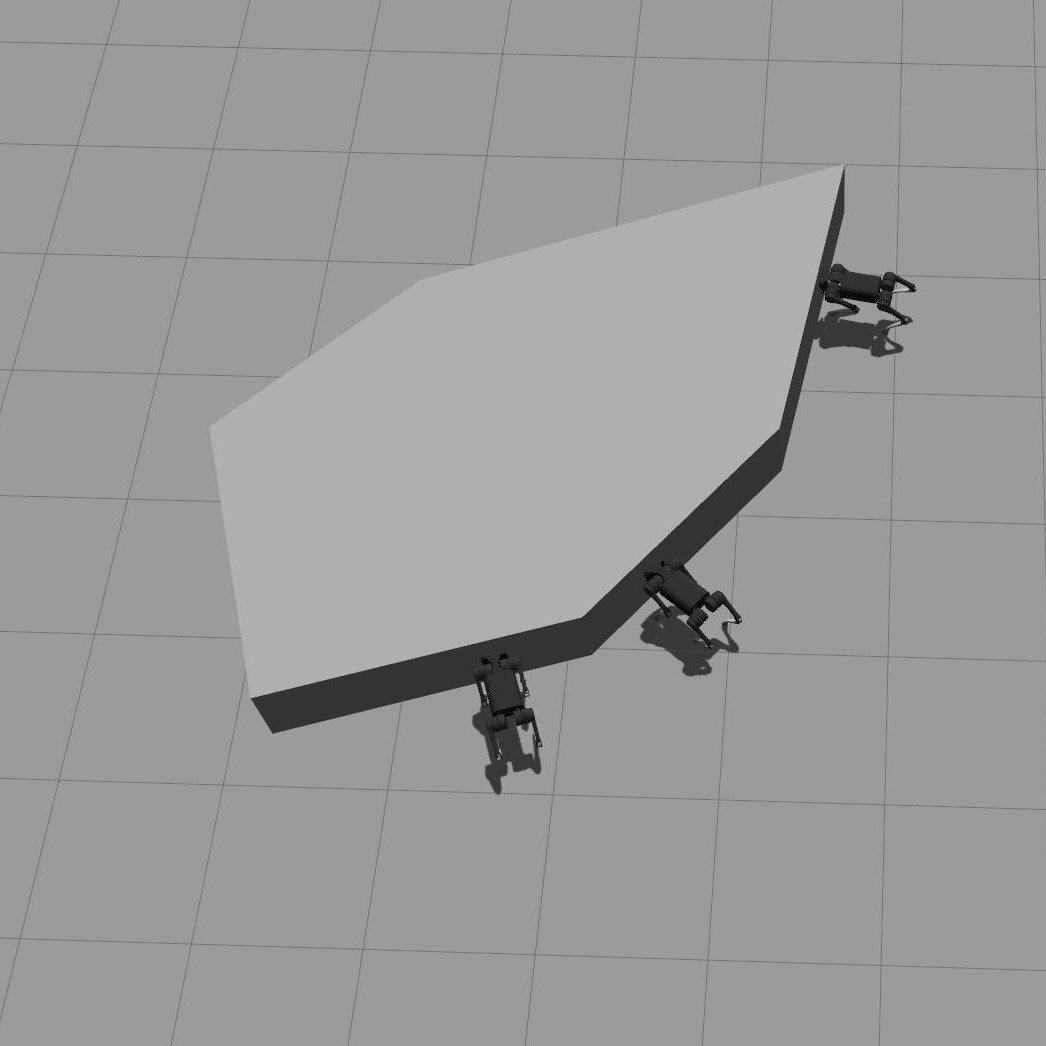}}
    \hfill
    \subfloat[Position error]{\includegraphics[width=1\linewidth]{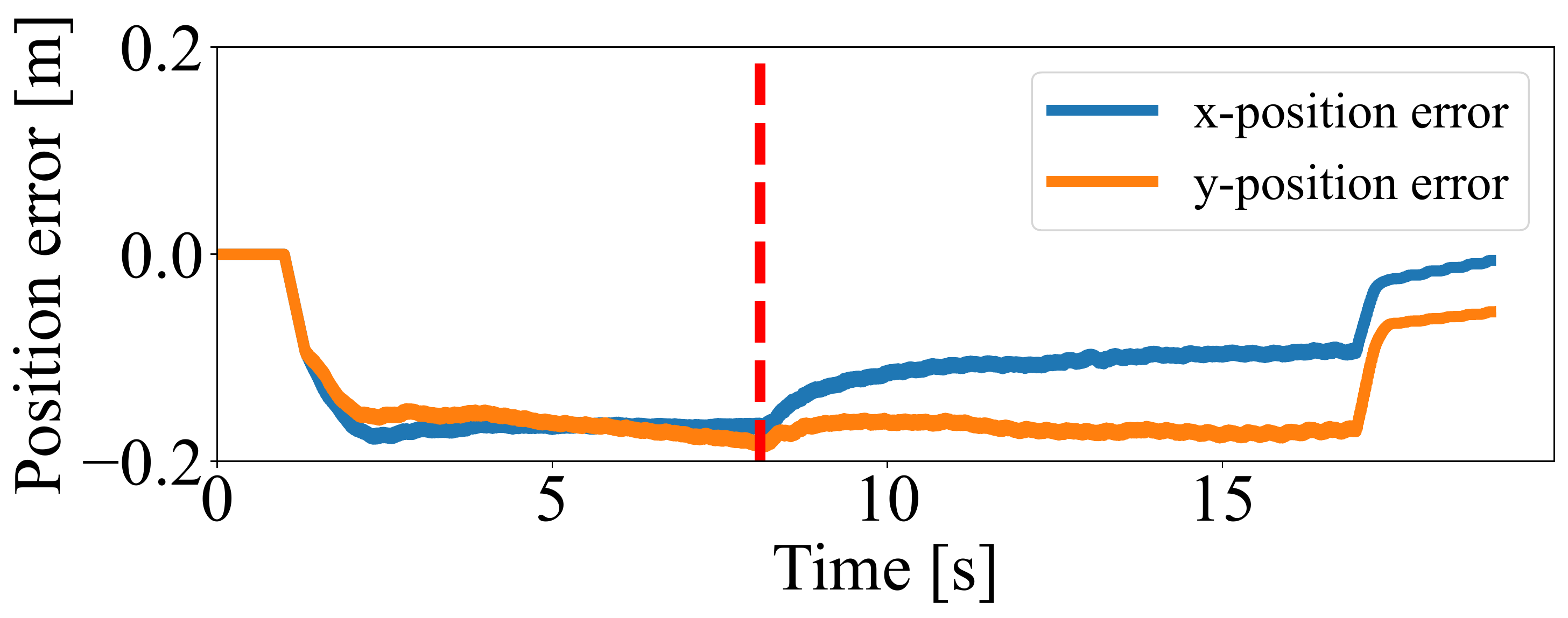}}
    \hfill
    \subfloat[Yaw tracking]{\includegraphics[width=1\linewidth]{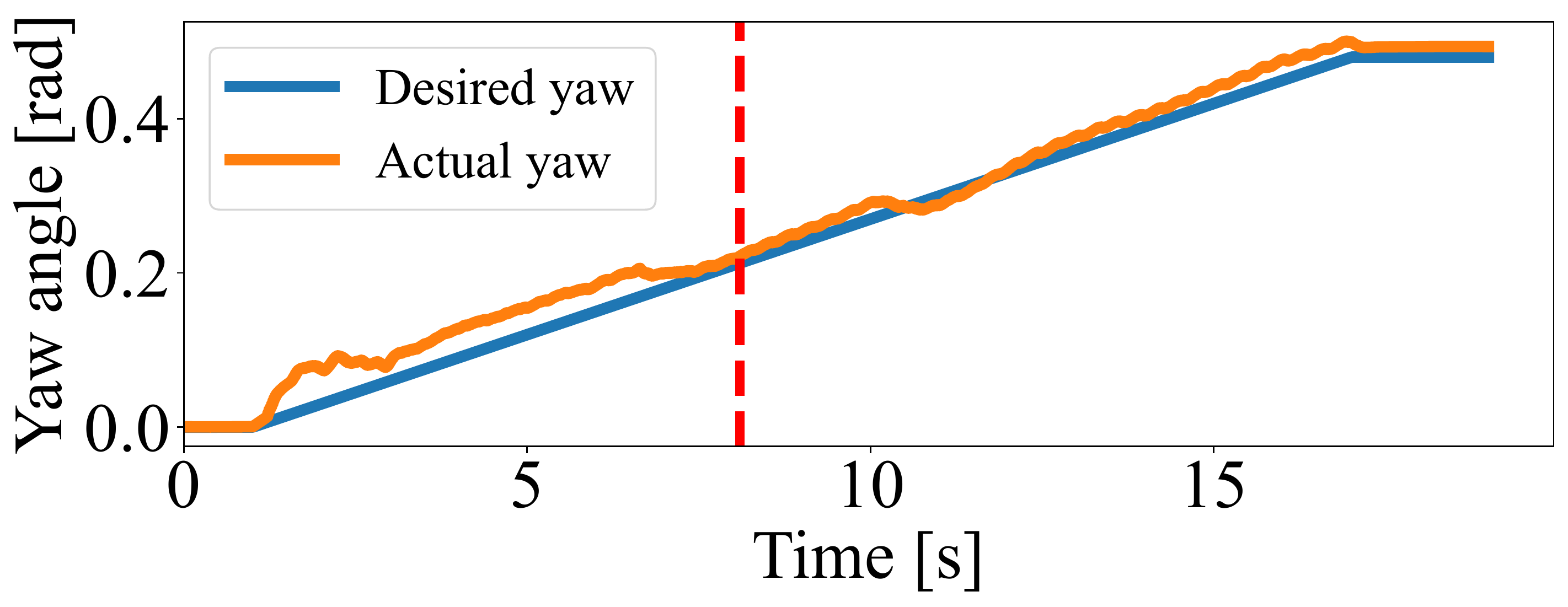}}

	\caption{{\bfseries{Collaborative manipulation for an unknown heavy 18 kg object}}. The manipulation task starts with two robots; then, the third robot joins them to get better performance. The red dashed line indicates the time the third robot joined the team.}
	\label{fig: three_robot}
		\vspace{-1em}
\end{figure}

Firstly, we intended to demonstrate the adaptability of our control system, even in the middle of an operation, to any number of robots. Secondly, we want to exhibit the effectiveness of our approach in manipulating heavy objects, which is unfeasible for a single robot to accomplish. To initiate the task, we employ two quadruped robots to manipulate an unknown heavy object weighing 18 kg. As the tracking error began to increase, we introduced another robot to the team to improve the tracking performance. By including the third robot, the QP-based controller system could distribute forces to all three robots, allowing them to collaborate optimally during the manipulation task. Therefore, the load on the other two robots, which has already reached its threshold, was reduced. The team's performance during this simulation is depicted in \figref{fig: three_robot}. Notably, the tracking error improved after the third robot joined the team, as indicated by the red dashed line.

\section{Conclusion} \label{sec: conclusion}
In summary, we propose a hierarchical adaptive control approach for the collaborative manipulation of a heavy, unknown object using a group of quadrupedal robots. The control framework comprises three levels. Firstly, an adaptive controller computes the manipulation force and moment. Secondly, a QP-based controller optimally distributes the force and moment among the robot team, as well as determines the optimal contact point for each robot. Finally, a decentralized loco-manipulation controller regulates the manipulation force of each robot while maintaining its stability. Our future work involves expanding the framework to a fully decentralized control system.

Our future objective is to implement this method in hardware experiments, where a team of robots will manipulate an unknown object through an obstacle-filled environment, effectively navigating around the obstacles.
\section*{Acknowledgments}
This work is supported in part by National Science Foundation Grant IIS-2133091. The opinions expressed are those of the authors and do not necessarily reflect the opinions of the sponsors.

\balance
\bibliography{references}

\end{document}